%% file: Template.tex
\documentclass{article}
\usepackage{spconf,amsmath,graphicx,hyperref}
\usepackage{soul,color}
\usepackage{caption}
\usepackage{diagbox}
\usepackage{amssymb}
\usepackage{textcomp}
\usepackage{multirow}
\usepackage{booktabs}
\usepackage{makecell}
\usepackage{bbding}
\usepackage{pifont}
\usepackage{xcolor}
\usepackage{colortbl}
\newcommand{\xmark}{\ding{55}}

\let\SUP

\makeatletter
\newcommand{\printfnsymbol}[1]{%
  \textsuperscript{\@fnsymbol{#1}}%
}
\makeatother

\title{Revisiting Modality Imbalance In Multimodal Pedestrian Detection}

\name{\parbox{0.85\linewidth}{\centering Arindam Das\textsuperscript{1,2}, Sudip Das\textsuperscript{1}, Ganesh Sistu\textsuperscript{3}, Jonathan Horgan\textsuperscript{3}, Ujjwal Bhattacharya\textsuperscript{4}, Edward Jones\textsuperscript{5}, Martin Glavin\textsuperscript{5}, and  Ciar\'{a}n Eising\textsuperscript{2,5}}}

\address{\textsuperscript{1}DSW, Valeo India, 
\textsuperscript{2}University of Limerick, Ireland, \textsuperscript{3}Valeo Vision Systems, Ireland,
\\
\textsuperscript{4}Indian Statistical Institute, Kolkata, \textsuperscript{5}University of Galway, Ireland\\
{\tt\small firstname.lastname@valeo.com, firstname.lastname@ul.ie}}

\let\oldtwocolumn\twocolumn
\renewcommand\twocolumn[1][]{
    \oldtwocolumn[{#1}{
    \begin{center}
            \captionsetup{singlelinecheck=false, font=small, belowskip=-6pt}
           \includegraphics[width=\textwidth]{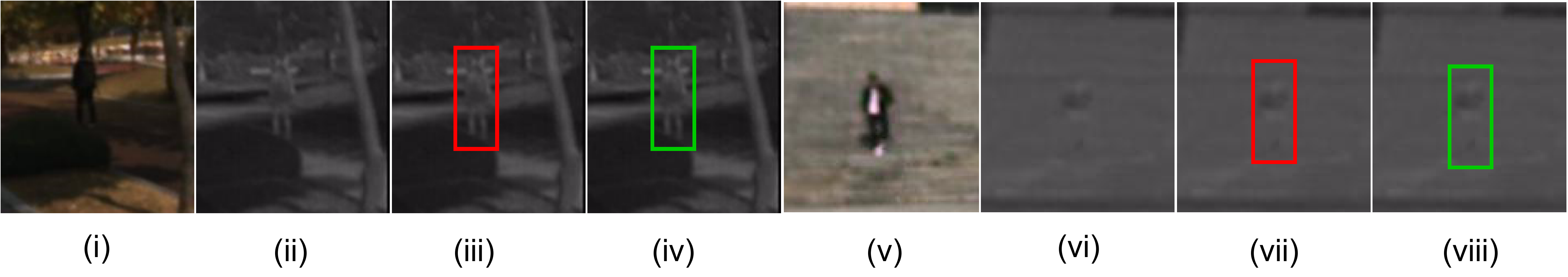}
           \captionof{figure}{\textbf{Results of superior modality guided multimodal feature fusion on KAIST \cite{hwang2015multispectral} dataset.} (i) Input RGB image where pedestrian is not recognizable due to low-light and dense shadow, however the same object is prominent in infrared (ii); (v) Input RGB image where pedestrian is far away from the camera but semantic features are rich in visible spectrum, but due to insignificant difference in heat between foreground and background, pedestrian features in thermal (vi) are not identifiable; (iii) and (vii) -  Detection results of the proposed method when the guidance from the superior modality is not accounted during multimodal feature fusion; (iv) and (viii) - Inference results when feature fusion is guided by the superior modality (infrared for the first example and RGB for the second one) via proposed regularizer to minimize modality imbalance. \textcolor{red}{Red} and \textcolor{green}{Green} colored bounding boxes represent missed and correct detections.}
           \label{fig:fig1}
        \end{center}
    }]
}

\begin{document}

\maketitle

\input{include/abstract.tex}

\begin{keywords}
Multimodal Learning, Modality Imbalance, Multimodal Feature Fusion, Pedestrian Detection.
\end{keywords}

\input{include/introduction.tex}

\input{include/architecture.tex}

\input{include/results.tex}

\input{include/conclusions.tex}

\bibliographystyle{IEEEbib}
\bibliography{strings,refs}

\end{document}

%% file: include/abstract.tex
\begin{abstract}
\vspace{1mm}
 
Multimodal learning, particularly for pedestrian detection, has recently received emphasis due to its capability to function equally well in several critical autonomous driving scenarios such as low-light, night-time, and adverse weather conditions. 
However, in most cases, the training distribution largely emphasizes the contribution of one specific input that makes the network biased towards one modality. 
Hence, the generalization of such models becomes a significant problem where the non-dominant input modality during training could be contributing more to the course of inference. 
Here, we introduce a novel training setup with regularizer in the multimodal architecture to resolve the problem of this disparity between the modalities. 
Specifically, our regularizer term helps to make the feature fusion method more robust by considering both the feature extractors equivalently important during the training to extract the multimodal distribution which is referred to as removing the imbalance problem. 
Furthermore, our decoupling concept of output stream helps the detection task by sharing the spatial sensitive information mutually.
Extensive experiments of the proposed method on KAIST and UTokyo datasets shows improvement of the respective state-of-the-art performance.

\end{abstract}

%% file: include/introduction.tex
\section{Introduction}
\label{sec:intro}
    \vspace{-4mm}
While cameras are one of the primary sensors for autonomous driving perception systems, they have generally failed in certain crucial scenarios. For example, 1) automotive cameras are generally mounted outside of the vehicle body that makes the camera lens directly exposed and high chance of becoming soiled \cite{das2019soildnet, das2020tiledsoilingnet} in the presence of sand, mud, dirt, snow, grass, etc.;
2) Sun glare \cite{yahiaoui2020let} obstructs downstream vision-based algorithms to work efficiently on the overexposed area; 3) the presence of dense shadow 
hinders especially the algorithms that operate at pixel level that include semantic segmentation, instance segmentation, etc; and 4) incorrect detection of pedestrians due to lack of information in camera data during low-light and night-time operation, leading to inaccurate pedestrian detection, and its extensions, such as the estimation of pedestrian pose \cite{das2022deep} \cite{kishore2019cluenet} \cite{das2020end}. Thus, in adverse conditions, the perception stack is not reliable enough for vehicle autonomy considering input only in the visible spectrum.
Therefore, it makes sense for autonomous driving systems to consider inputs from different modalities, such as depth \cite{rashed2018depth}, thermal \cite{dasgupta2022spatio}, LiDAR \cite{el2019rgb}, and others, in order to maintain the high safety requirements of vehicle automation. In this paper, we specifically look at a multimodal learning approach to detection using visual and thermal data.
 
\begin{figure*}[!ht]
    \centering
    \includegraphics[width=\textwidth]{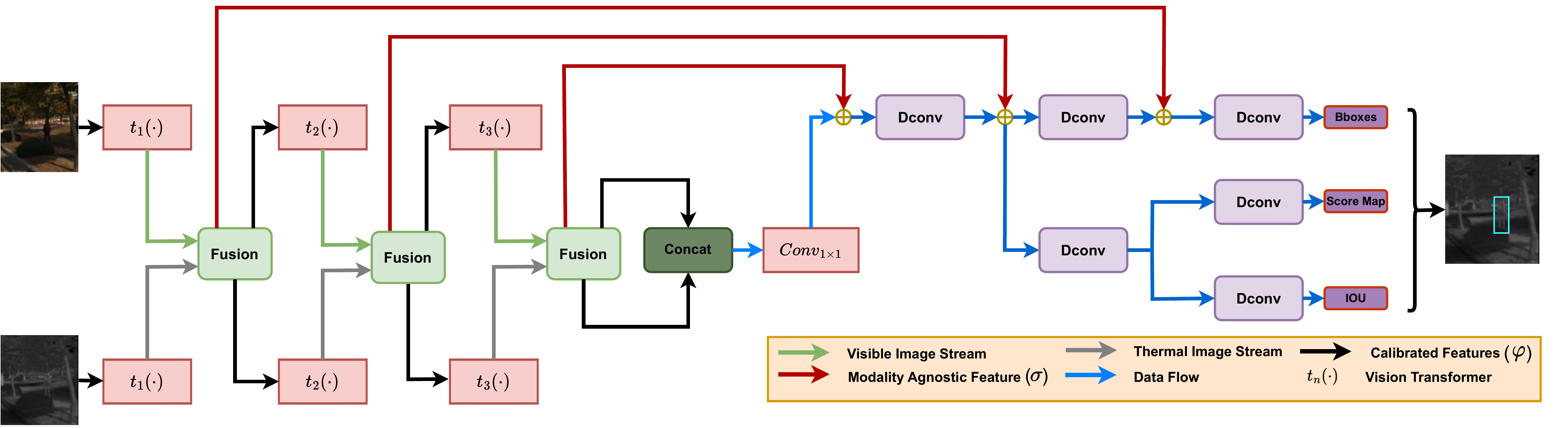}
    \vspace{-.27in}
    \caption{Our proposed End-to-End learning framework for multimodal pedestrian detection in diverse scenarios.}
    \label{fig:architecture}
    \vspace{-5mm}
\end{figure*}

When considering multimodal learning, scenes in which one sensor performs significantly better than the other sensor type can cause a bias in training towards one modality. For example, if we train with only scenes from night-time, then there will be a bias towards thermal, limiting the generalisability of the network.
The goal, therefore, is to bring a less imbalanced multimodal training scheme by reducing such biases, with the aim for multimodal feature fusion to become robust. Zhou \textit{et al.} \cite{zhou2020improving} addressed the problem of modality imbalance from two different aspects - illumination and feature. Layer-wise fusion was proposed between two unimodal stream networks through Differential Modality Aware Fusion (DMAF) module.
However, the reported solution is based on architectural design that makes it difficult to regularize any modality. 
Likewise, in \cite{dasgupta2022spatio}, the fine-tuned fused features are further re-calibrated equally for both modalities. 

There is, therefore, potential for further improvement in regularization between modalities without the need for any extra inputs to the network. The main contributions of our work are as follows. 1) We introduce a new concept to balance the information of both modalities by leveraging Logarithmic Sobolev Inequality to equivalently consider the information during the multimodal fusion; 2) We propose a novel end-to-end multi-modal network to achieve improvements over the recent state-of-the-art methods on two publicly available benchmark datasets, KAIST\cite{hwang2015multispectral} and UTokyo \cite{takumi2017multispectral}; and, 3) We conduct a systematic ablation study involving different backbones, training strategies, and network components.

%% file: include/architecture.tex
\section{Proposed Approach}

The proposed end-to-end architecture, illustrated in Figure \ref{fig:architecture}, combines several components that are discussed below.
\subsection{Multimodal Feature Extraction and Fusion}

Recently, the Pyramid Vision Transformer (PVT) \cite{wang2021pyramid} has shown exemplary performance in dense prediction tasks, especially for smaller objects, which makes PVT a suitable feature extractor in this work. 
We employed two instances of PVT specific to visible and thermal inputs - $E_{V}(.)$ and $E_{T}(.)$. 
We have added the same feature fusion unit as used in \cite{dasgupta2022spatio}, where the modality agnostic raw feature vector ($\sigma$) is passed and each element of the vector is multiplied with each channel before respective deconvolution layers, as shown in the proposed architecture diagram (Figure \ref{fig:architecture}).
We concatenate the calibrated features ($\varphi$) of both the streams, passing them through a $Conv_{1 \times 1}$ to reduce the dimension similar to the last layer of the encoder. We represent the latent variables which are an inner representation of the encoding model to use  further for the downstream tasks.   

\subsection{Multi-stream Decoupled Detection Branch}

In most of the existing methods, such as \cite{dasgupta2022spatio}, final predictions are extracted from a single-stage detection decoder. In this work, we combine Score Map and IoU as outputs from one branch and decouple bounding box output to a separate branch. The main reason for creating a multi-stream decoupled detection branch is to group the related tasks. In multitask learning setup, pixel-wise classification helps with the detection task for estimating the bounding box (BBoxes) of the pedestrians accurately in different scales and various occlusion scenarios. To achieve this, our decoupled output streams are designed in such a way that IoU map further helps the task of the score map to improve the performance while the region of the objects in the score map is shifted.
Both streams consist of multiple deconvolution layers, where layers are progressively upsampled by a factor of $2$ and the number of channels is reduced by half. 
Class balanced cross-entropy loss and IoU loss, as used in \cite{dasgupta2022spatio}, are  applied for Score Map and IoU regression.
Repulsion loss \cite{wang2018repulsion}, which was proposed exclusively to handle crowded pedestrian scenarios, is applied for the BBoxes regression.

\subsection{Detection Loss Function}

We developed a training setup with Logarithmic Sobolev Inequalities \cite{gross1975logarithmic} to deal with the problem of training a network with multimodal data that allows the extraction of equivalent information from both modalities. 
Here, we consider the visible and thermal images as multimodal inputs for the training of a multimodal model in the application of pedestrian detection. 
The training data $X$ (i.e., $x_{t}^{k}$ and $x_{v}^{k}$, for thermal and visual respectively) where each of the instance $x^{i}$ that consists of visible image $x_{v}^{k}$ and thermal image $x_{t}^{k}$, with their corresponding ground truth $y^{k}$, and $\,m^{k}$ represents the multimodal features in the probability measure space. 
The measure of the uncertainty of the random variables $H(\cdot)$ via Softmax Entropy between the distributions of $g(\hat{y})$ and $f_{w}(\hat{y})$ is:
 \begin{equation} \label{eqn_entrphy}
	H(g, f_{w}) = - \sum_{\hat{y}} g(\hat{y}) \log f_{w} (\hat{y} | x_{t}^{k}, x_{v}^{k})
\end{equation}
Throughout, we consider a transformation function $f_w(\cdot)$, parameterized by $w$, and $g(\hat{y})$ is the ground truth. Additionally, we define $w_v \in W$ and $w_t \in W$ to denote the parameters for visible stream and thermal stream respectively.

Through the optimization process of the parameters, it is possible that the network trains more on a single modality due to one specific input scheme being emphasized as a result of different lighting conditions, weather conditions, etc. Hence, an approximation of an imbalance function $f_{w}(\hat{y})$ towards a single modality leads to poor fusion. To resolve the problem of multimodal inequality, we apply Fisher information that helps to measure the amount of \textit{modality-specific} information in each input distribution (i.e., $x_{t}^{k}$ and $x_{v}^{k}$) to train the parameters $w_t$ and $w_v$ respectively as described in (\ref{eqn_dlsi}).

We consider the Logarithmic Sobolev method that is encapsulated for extracting the equally important features from multimodal distribution.
 \begin{equation} \label{eqn_LSI}
    \begin{split}
	\int_{R^{n}} \| f(X)\|^{2} \log{ \| f(X) \| } \, du^{X}(x) \\ \leq \int_{R^{n}} \|  \nabla  
 f(X) \|^{2} \,du^{X}(x) + \lVert f(X) \rVert_{2}^{2} \log \lvert f(X) \rvert_{2}
    \end{split} 
\end{equation}
where $\,du^{X}(X)$ is the probability density function and $u$ denotes the Gaussian measure on $\,R^{2}$. $\lVert f(\cdot)\rVert$ is the norm on  the Hilbert space $\, L^{2}$. We derive the following equation where the function $f(x) \geq 0 $, 
\begin{equation} \label{eqn_dlsi}
    \begin{split}
    \int_{R^{n}} f(X) \log f(X) \, du^{X}(X) - \\ \int_{R^{n}} f(X) \, du^{X}(X) \log \int_{R^{n}} f(X) du^{X}(X) \\ \leq \frac{1}{2} \int_{R^{n}} f(X) \frac{\| \nabla f(X) \| ^{2}}{f(X)} \, du^{X}(x)
    \end{split}
 \end{equation}
The above equation describes that entropy is non-negative function since the formulation of Fisher information is non-negative. It also bounds the functional entropy using the method of Fisher information through the log Sobolev inequality. The functional entropy $E(f(X))$ is as follows,
\begin{equation} \label{entr_eqn_dlsi}
    \begin{split}
    E(f(X)) \cong  \int_{R^{n}} f(X) \log f(X) \, du^{X}(X) - \\ \int_{R^{n}} f(X) \, du^{X}(X) \log \int_{R^{n}} f(X) du^{X}(X) 
    \end{split}
 \end{equation}
 
In the problem of maximizing the information in latent space, we denote measures $\,u^{x_{t}}$ and $\,u^{x_{v}}$ respectively for visible and thermal feature distributions. During optimization, the measures $\,u^{x_{t}}$ and $\,u^{x_{v}}$ are in Gaussian distribution and denoted as $ \,u^{x_{t}} \sim \mathcal{N}(\mu_{x_{t}},\,\sigma_{x_{t}}^{2})$ and $ \,u^{x_{v}} \sim \mathcal{N}(\mu_{x_{v}},\,\sigma_{x_{v}}^{2})$. 
Here,  ($\mu_{x_{t}}$, $\mu_{x_{v}}$) and ($\sigma_{x_{t}}^{2}$, $\sigma_{x_{v}}^{2}$)  are the mean and variance of the measures of the multimodal features. 
The product measure over both distributions for (\ref{eqn_dlsi}) is defined as $ \,u^{X} = \,u^{x_{t}} \bigotimes \,u^{x_{v}}$. 
We define a function $S^X(\cdot)$ in (\ref{entropy}) to calculate the sensitivity of the softmax function
$p_{w}( \cdot | x_{v}, x_{t})$ 
for our proposed architecture to the Gaussian measures 
$u^{x_{v}}$ and $u^{x_{t}}$.
\begin{equation} \label{entropy}
S^X(\cdot) =  H(p_{w}( \cdot | u^{x_{v}}, u^{x_{t}}), p_{w}(\cdot | x_{v}, x_{t}))
\end{equation}
Hence, we replace the sensitivity function in (\ref{eqn_dlsi}) with the following regularization,
\begin{equation} \label{sensitivity}
\lambda_{regu} =  \int_{R^{n}} \frac{\| \nabla S^X(\cdot) \| ^{2}}{S^X(\cdot)} \, du^{X}(X)
\end{equation}
The regularization $\lambda_{regu}$ is applied with the loss function of score map,  denoted as $\lambda_{bce_{regu}}$.
\begin{equation} \label{balance_regu}
\lambda_{bce_{regu}} =  \beta \lambda_{bce} + \delta \lambda_{regu}
\end{equation}

The minimization of the cost function to estimate $f_w(\cdot)$ is effective since the detector never saturates unless and until it detects the pedestrians precisely with high probability by leveraging the features, importantly, from both modalities.

Our final cost function is made up of four components,
\begin{equation} \label{final_lossfunction}
    \lambda_{loss} = \alpha \lambda_{rep} + \beta \lambda_{bce} + \delta \lambda_{regu} + \gamma \lambda_{iou}
 \end{equation}
where $\lambda_{rep}$ is the repulsion loss \cite{wang2018repulsion} to minimize the error between the predicted and ground truth boxes and also to accurately fit the predicted bounding boxes for occluded pedestrians in crowded scenarios. 
We make use of binary cross entropy loss \cite{dasgupta2022spatio} denoted as $\lambda_{bce}$ to calculate the error of the score map. 
Additionally, we use $\lambda_{iou}$ to over-penalize the overlapping error on the detected object and ground truth for precision. 
Both tasks mutually share the gradient to the early layers of the detection network. $\alpha$, $\beta$, $\gamma$, and $\delta$ are the hyperparameters to balance the four different auxiliary losses.

%% file: include/results.tex
\newcommand{\etal}{\textit{et al.}}

\section{Experimentation Details}
\subsection{Datasets and Training Details}
KAIST \cite{hwang2015multispectral} is a popular multimodal dataset used for pedestrian detection tasks. It contains approximately $95,000$ pairs of \textit{color-thermal} frames with a total of $1,182$ unique pedestrians and $103,128$ annotated bounding boxes. Zhang \textit{et al.} \cite{zhang2019weakly} fixed the alignment issues of this dataset and Liu \textit{et al.} \cite{liu2016multispectral_} published the refined annotations for the test set. The latest version of this dataset with these fixes are used in this work. UTokyo \cite{takumi2017multispectral} dataset consists of $7,512$ frames captured using RGB, far-infrared (FIR), mid-infrared (MIR), and near-infrared (NIR) cameras, are used in this work for cross dataset generalization. We replicated the training strategy discussed in \cite{dasgupta2022spatio}.

\begin{figure}
    \centering
    \includegraphics[height=.87in]{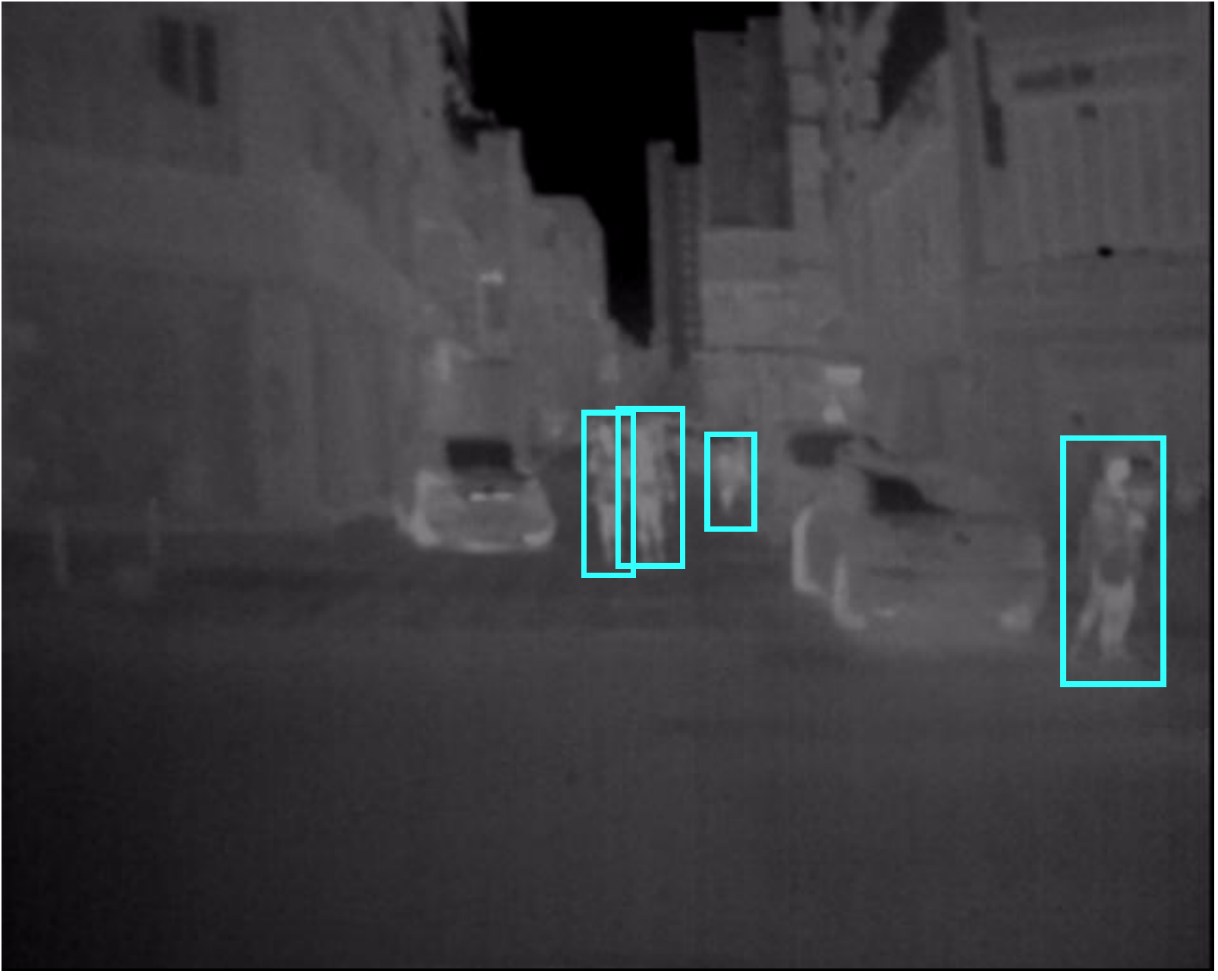}
    \includegraphics[height=.87in]{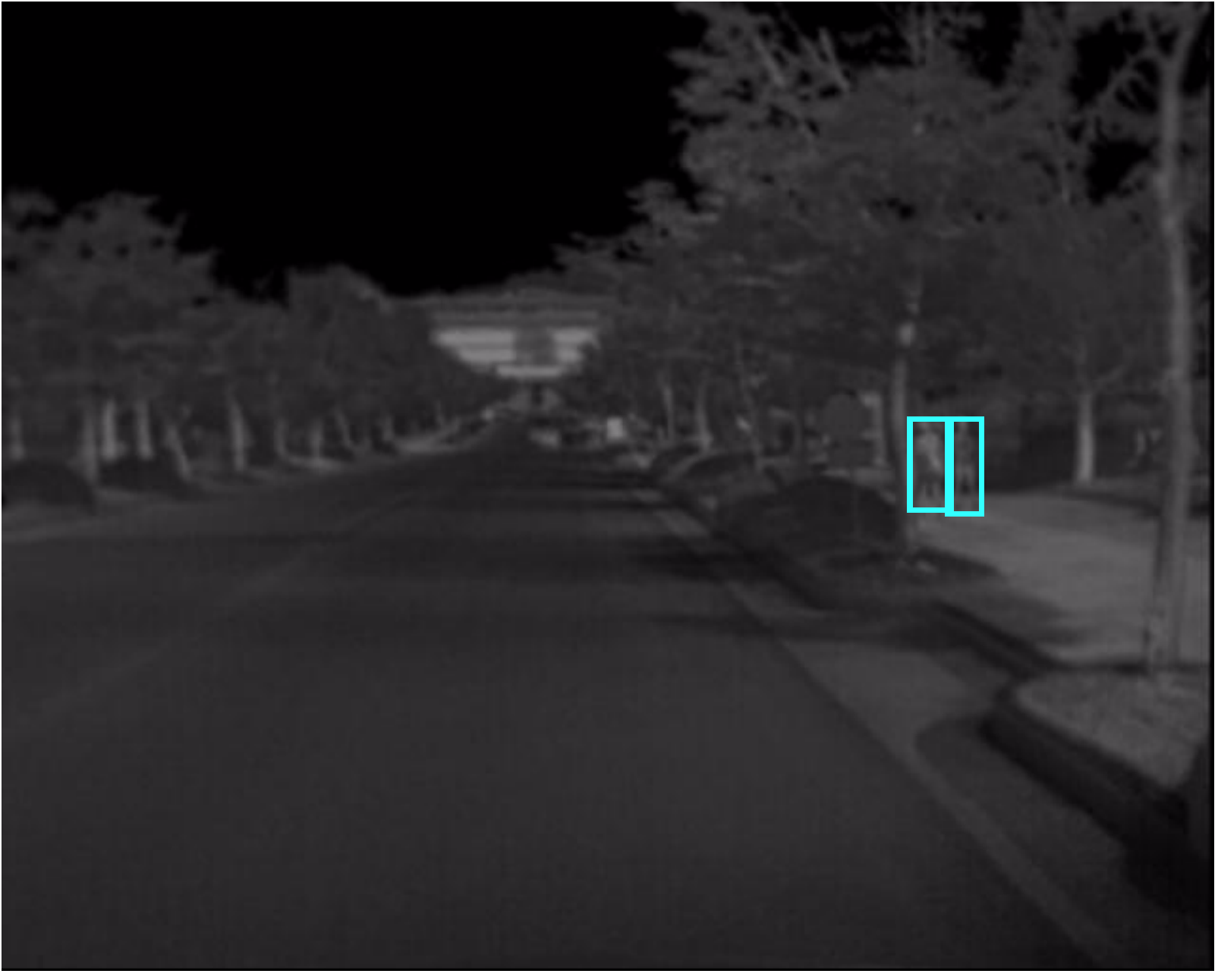}
     \includegraphics[height=.87in]{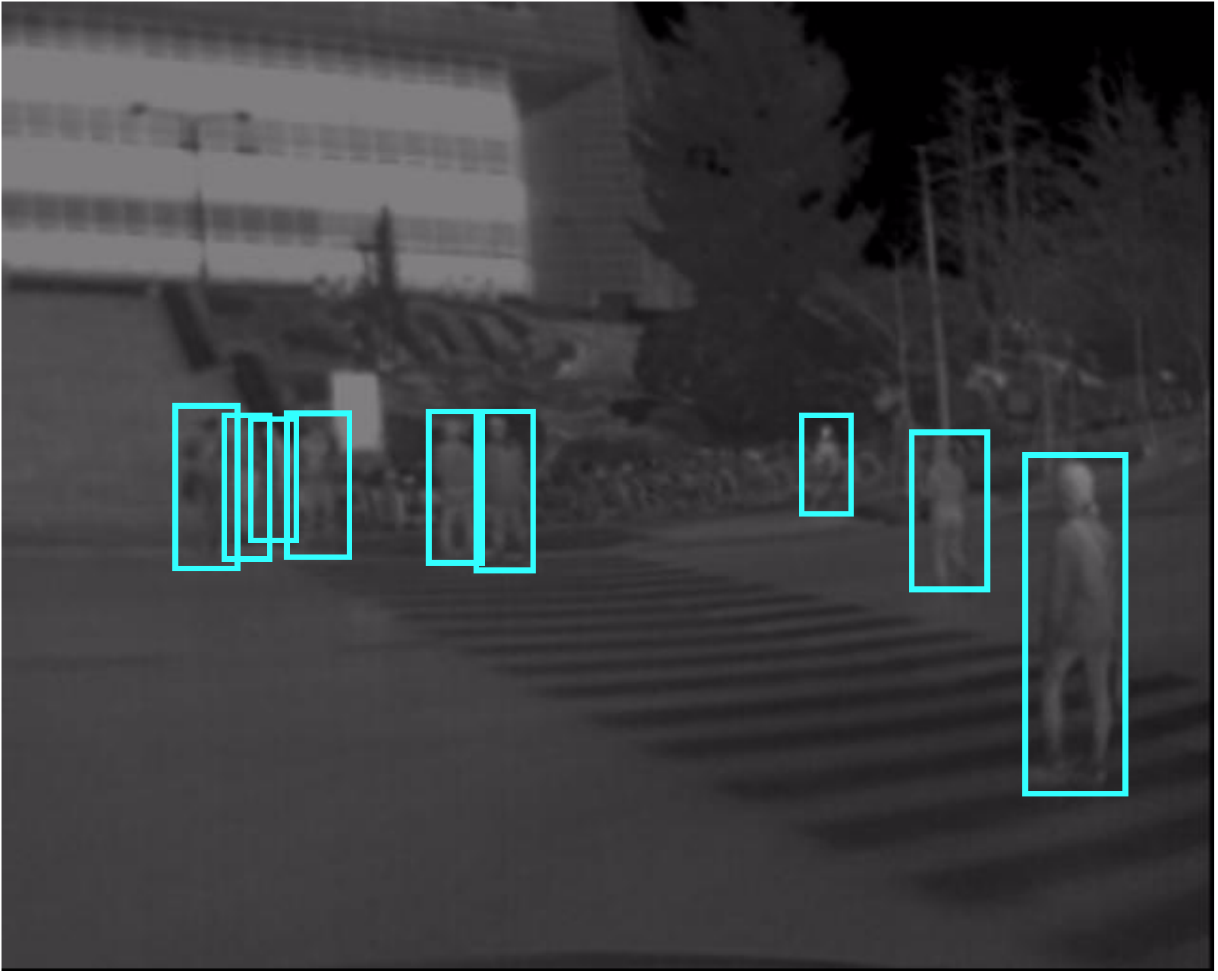}
    \caption{Inference results of the proposed method on the KAIST \cite{hwang2015multispectral} dataset.}
    \vspace{5mm}
    \label{fig:experimental_results}
\end{figure}

\begin{table}[!h]
\centering
    \scalebox{.64}{
    \begin{tabular}{lcccccccccc}
    \toprule
        \textbf{Backbone} & \textbf{C.Learning} & \textbf{Regularizer} & \multicolumn{4}{c}{\textbf{No. of Fusion Units}} &   \multicolumn{3}{c}{\textbf{Miss Rate} $\downarrow$} \\
        \cmidrule(r){4-7}
        \cmidrule(r){8-10}
        & & & \textbf{1} & \textbf{2}  & \textbf{3} & \textbf{4} & \textbf{All} & \textbf{Day} & \textbf{Night}\\
         \cmidrule(r){1-10} 
         
    \multirow{19}{*}{PVT \cite{wang2021pyramid}} & \color{red}\xmark& \color{red}\xmark &  \color{green}\Checkmark & \color{red}\xmark  & \color{red}\xmark & \color{red}\xmark & 28.16 & 29.72 & 27.39 \\

& \color{red}\xmark & \color{red}\xmark & \color{red}\xmark & \color{green}\Checkmark & \color{red}\xmark & \color{red}\xmark & 27.27 & 27.01 &  26.16\\
   
& \color{red}\xmark & \color{red}\xmark & \color{red}\xmark & \color{red}\xmark & \color{green}\Checkmark & \color{red}\xmark & 25.62 &  26.92 & 24.62 \\

& \color{red}\xmark & \color{red}\xmark & \color{red}\xmark & \color{red}\xmark & \color{red}\xmark & \color{green}\Checkmark  & 25.38 &  26.76 & 24.66 \\
   
  \cmidrule(r){2-10}

& \color{green}\Checkmark & \color{red}\xmark & \color{green}\Checkmark  & \color{red}\xmark  & \color{red}\xmark & \color{red}\xmark & 27.23 & 28.12 & 26.87 \\

& \color{green}\Checkmark & \color{red}\xmark & \color{red}\xmark & \color{green}\Checkmark & \color{red}\xmark & \color{red}\xmark & 26.18  & 25.39 & 25.68 \\
   
& \color{green}\Checkmark & \color{red}\xmark & \color{red}\xmark & \color{red}\xmark & \color{green}\Checkmark & \color{red}\xmark  & 24.47 & 25.32 & 23.81 \\

& \color{green}\Checkmark & \color{red}\xmark & \color{red}\xmark & \color{red}\xmark & \color{red}\xmark & \color{green}\Checkmark   & 24.87 & 26.2 & 24.19 \\

  \cmidrule(r){2-10}

& \color{red}\xmark  & \color{green}\Checkmark & \color{green}\Checkmark  & \color{red}\xmark  & \color{red}\xmark & \color{red}\xmark & 22.53 & 23.29   & 21.2 \\
   
& \color{red}\xmark  & \color{green}\Checkmark & \color{red}\xmark & \color{green}\Checkmark & \color{red}\xmark & \color{red}\xmark & 21.02  & 22.13 & 21.38 \\
   
& \color{red}\xmark  & \color{green}\Checkmark & \color{red}\xmark & \color{red}\xmark & \color{green}\Checkmark & \color{red}\xmark  & 17.77 & 18.18 & 17.31 \\

& \color{red}\xmark  & \color{green}\Checkmark & \color{red}\xmark & \color{red}\xmark & \color{red}\xmark & \color{green}\Checkmark   & 16.46 & 17.79 & 16.28 \\

 \cmidrule(r){2-10}

& \color{green}\Checkmark  & \color{green}\Checkmark & \color{green}\Checkmark  & \color{red}\xmark  & \color{red}\xmark & \color{red}\xmark & 17.51 & 18.68   & 17.94 \\
   
& \color{green}\Checkmark  & \color{green}\Checkmark & \color{red}\xmark & \color{green}\Checkmark & \color{red}\xmark & \color{red}\xmark & 11.31  & 12.99 & 11.78 \\
   
& \color{green}\Checkmark  & \color{green}\Checkmark & \color{red}\xmark & \color{red}\xmark & \color{green}\Checkmark & \color{red}\xmark  & \textbf{7.41} & \textbf{7.69} & \textbf{7.03} \\

& \color{green}\Checkmark  & \color{green}\Checkmark & \color{red}\xmark & \color{red}\xmark & \color{red}\xmark & \color{green}\Checkmark   & 8.43 & 8.26 & 8.97 \\
    
    \bottomrule
    \end{tabular}}

    \caption{Ablation study of the proposed architecture on KAIST dataset \cite{hwang2015multispectral}. } 
    \vspace{-7mm}
    \label{tab:arch_ablation}
\end{table}

\subsection{Ablation Study}
We considered KAIST \cite{hwang2015multispectral} to perform our ablation study as it is quite popular and contains large number of samples. The results of the ablation study are reported on the test set using standard log average Miss Rate (MR) to estimate the error. Due to the recent improvement shown by PVT \cite{wang2021pyramid} for smaller objects, we designed a series of experiments where we kept PVT as a baseline. This ablation study includes our proposed regularizer that is enabled and disabled with all possible combinations of network components to validate our proposal. Number of instances of feature fusion units from \cite{dasgupta2022spatio} are progressively added in the network from $1$ to $4$. Curriculum Learning \cite{bengio2009curriculum} was considered it has been helpful to achieve better generalization in our previous works. Table \ref{tab:arch_ablation} summarizes the results of the ablation study and indicates the optimal configuration of the network and training strategy. Further we performed ablation study on different backbones using the optimal configuration, reported in Table \ref{tab:abliation_encoder} where PVT has outperformed other encoders by large margin.

\begin{table}[!h]
\centering

\scalebox{.7}{
\begin{tabular}{|c|c|c|c|}
\hline
\textbf{Backbone} & \textbf{MR(ALL)}$\downarrow$ & \textbf{MR(Day)}$\downarrow$ & \textbf{MR(Night)}$\downarrow$\\ \hline \hline 
ResNet-50   \cite{he2016deep}   &  15.57  &   15.89     &  15.1      \\ \hline
ResNet-101  \cite{he2016deep}    &  12.4  &   12.8     &  11.1      \\ \hline
ResNeXt-101 \cite{xie2017aggregated}     &  11.73  &  11.9  &  11.47       \\ \hline
PVT \cite{wang2021pyramid}     &  \textbf{7.41}  &  \textbf{7.69}  &  \textbf{7.03}  \\ \hline
\end{tabular}}
\caption{Ablation study of different backbone architectures
on KAIST dataset.}
    \vspace{3mm}
\label{tab:abliation_encoder}
\end{table}

\subsection{Evaluation results}

To facilitate fair visual comparison and verify our proposed regularizer, we have performed inference on a few samples where the network is trained with and without our proposed modified Logarithmic Sobolev Inequalities. In Figure \ref{fig:fig1}, it can be clearly observed when the multimodal learning is not accompanied by regularizing the input schemes based on the training distribution then the network tends to miss the detection when the representation of the same object is severely poor in one of the modalities. Table \ref{tab:KAIST_results} and \ref{tab:UTokyo_results} compares the proposed method with other existing state-of-the-art approaches using MR on KAIST and UTokyo respectively. We obtain incremental improvement in all categories for both datasets. All the experiments performed on KAIST \cite{hwang2015multispectral} dataset was evaluated as per the \textit{reasonable} setup \cite{dollar2012pedestrian} protocol.

\begin{table}[!h]
\centering

\scalebox{.7}{
\begin{tabular}{|c|c|c|c|}
\hline
\textbf{Architecture} & \textbf{MR(ALL)}$\downarrow$ & \textbf{MR(Day)}$\downarrow$ & \textbf{MR(Night)}$\downarrow$\\ \hline \hline 
MSDS-RCNN  \cite{li2018multispectral}    &  11.63  &   10.60     &  13.73       \\ \hline
CS-RCNN   \cite{zhang2020attention}   &  11.43  &   11.86     &  8.82      \\ \hline
AR-CNN  \cite{zhang2019weakly}    &  9.34  &   9.94     &  8.38      \\ \hline
MBNet \cite{zhou2020improving}     &  8.13  &  8.28  &  7.86       \\ \hline
Dasgupta \etal \cite{dasgupta2022spatio}     &  8.07  &  8.16  &  7.51  \\ \hline
\textbf{Ours}  &  \textbf{7.41}  &  \textbf{7.69}  &  \textbf{7.03}  \\ \hline
\end{tabular}}
\caption{Quantitative comparison of pedestrian detection on KAIST \cite{hwang2015multispectral} dataset.}
    \vspace{-5mm}
\label{tab:KAIST_results}
\end{table}

\begin{table}[!h]
\centering

\scalebox{.7}{
\begin{tabular}{|c|c|c|c|}
\hline
\textbf{Architecture} & \textbf{MR(ALL)}$\downarrow$ & \textbf{MR(Day)}$\downarrow$ & \textbf{MR(Night)}$\downarrow$\\ \hline \hline 
Halfway Fusion  \cite{li2018multispectral}    &  37.0  &   38.1     &  34.4       \\ \hline
Park et al.  \cite{park2018unified}   &  31.4  &   31.8     &  30.8      \\ \hline
AR-CNN \cite{zhang2019weakly}    &  22.1  &   24.7     &  18.1      \\ \hline
MBNet \cite{zhou2020improving}     &  21.1  &  24.7  &  13.5       \\ \hline
Dasgupta \etal \cite{dasgupta2022spatio}     & 19.04  &  20.32  &  12.86  \\ \hline

\textbf{Ours}      &  \textbf{17.29}  &  \textbf{18.73}  &  \textbf{10.09}  \\ \hline

\end{tabular}}
\caption{Quantitative comparison of pedestrian detection on  UTokyo \cite{takumi2017multispectral} dataset.}
    \vspace{-7mm}
\label{tab:UTokyo_results}
\end{table}

%% file: include/conclusions.tex
\section{Conclusions}
In this work, we introduced a novel end-to-end multimodal architecture and a cost-function regularizer that enables the fusion of features based on the contributions made among the modalities. Our independent output nodes are designed to decouple outputs that outperform the other coupled output stream variants. Our comparative analysis establishes the success of the proposed regularizer to reduce the modality imbalance in the network achieving accurate detection of pedestrians using multimodal data, thus obtaining state-of-the-art results on two public datasets - KAIST and UTokyo. In future work, we plan to revisit the problem of modality imbalance in a multi-task learning scenario.